\documentclass[conference]{IEEEtran}
\IEEEoverridecommandlockouts
\usepackage{cite}
\usepackage{float}
\usepackage{amsmath,amssymb,amsfonts}
\usepackage{algorithmic}
\usepackage{graphicx}
\usepackage{textcomp}
\usepackage{xcolor}
\usepackage{graphicx,subfigure}
\usepackage{multirow}
\usepackage{float}
\usepackage{url}

\def\BibTeX{{\rm B\kern-.05em{\sc i\kern-.025em b}\kern-.08em
    T\kern-.1667em\lower.7ex\hbox{E}\kern-.125emX}}
\begin{document}

\title{Sensor-invariant Fingerprint ROI Segmentation Using Recurrent Adversarial Learning}

\author{\IEEEauthorblockN{Indu Joshi \hspace{0.5cm}}
\IEEEauthorblockA{
\textit{IIT Delhi}, India\\
indu.joshi@cse.iitd.ac.in}
\and
\IEEEauthorblockN{Ayush Utkarsh$^{\star}$\thanks{$\star$ Equal contribution from both authors}}
\IEEEauthorblockA{
\textit{Independent Researcher}, India\\
ayushutkarsh@gmail.com}
\and
\IEEEauthorblockN{Riya Kothari$^{\star}$}
\IEEEauthorblockA{
\textit{USC}, USA\\
rskothar@usc.edu}
\and
\IEEEauthorblockN{Vinod K Kurmi}
\IEEEauthorblockA{
\textit{IIT Kanpur}, India\\
vinodkk@iitk.ac.in}
\and
\hspace{2cm}\IEEEauthorblockN{Antitza Dantcheva}
\IEEEauthorblockA{
\hspace{2cm}\textit{Inria Sophia Antipolis}, France\\
\hspace{2cm}antitza.dantcheva@inria.fr}
\and
\IEEEauthorblockN{Sumantra Dutta Roy}
\IEEEauthorblockA{
\textit{IIT Delhi}, India\\
sumantra@ee.iitd.ac.in}
\and
\IEEEauthorblockN{Prem Kumar Kalra}
\IEEEauthorblockA{
\textit{IIT Delhi}, India\\
pkalra@cse.iitd.ac.in}
}

\maketitle

\begin{abstract}
A fingerprint region of interest (roi) segmentation algorithm is designed to separate the foreground fingerprint from the background noise. All the learning based state-of-the-art fingerprint roi segmentation algorithms proposed in the literature are benchmarked on scenarios when both training and testing databases consist of fingerprint images acquired from the same sensors. However, when testing is conducted on a different sensor, the segmentation performance obtained is often unsatisfactory. As a result, every time a new fingerprint sensor is used for testing, the fingerprint roi segmentation model needs to be re-trained with the fingerprint image acquired from the new sensor and its corresponding manually marked ROI. Manually marking fingerprint ROI is expensive because firstly, it is time consuming and more importantly, requires domain expertise. In order to save the human effort in generating annotations required by state-of-the-art, we propose a fingerprint roi segmentation model which aligns the features of fingerprint images derived from the unseen sensor such that they are similar to the ones obtained from the fingerprints whose ground truth roi masks are available for training. Specifically, we propose a recurrent adversarial learning based feature alignment network that helps the fingerprint roi segmentation model to learn sensor-invariant features. Consequently, sensor-invariant features learnt by the proposed roi segmentation model help it to achieve improved segmentation performance on fingerprints acquired from the new sensor. Experiments on publicly available FVC databases demonstrate the efficacy of the proposed work.
\end{abstract}


\section{Introduction}
Biometric-based authentication systems are used for a gamut of applications such as law enforcement, border security, surveillance, etc. Among all the different biometric modalities which may require expensive sensing device or may not work reliably in uncontrolled settings, fingerprint is one of the most widely used modality. One of the key component which attributes to the robustness of a fingerprint matching system is the fingerprint region of interest (roi) segmentation module. A fingerprint roi segmentation module is dedicated towards segmenting foreground fingerprint region with clear ridge patterns from the background noise. Noise in a fingerprint can  originate due to presence of oil, grease or dirt on surface of the fingerprint sensor used to acquire the fingerprints. A fingerprint roi segmentation module serves dual purpose. Firstly, it minimizes spurious minutiae (feature) detection which translates to improved matching performance. Secondly, it limits the matching to only foreground which reduces computational time for matching.
\begin{figure}
\centering
\includegraphics[width=8.5cm,height=10.5cm]{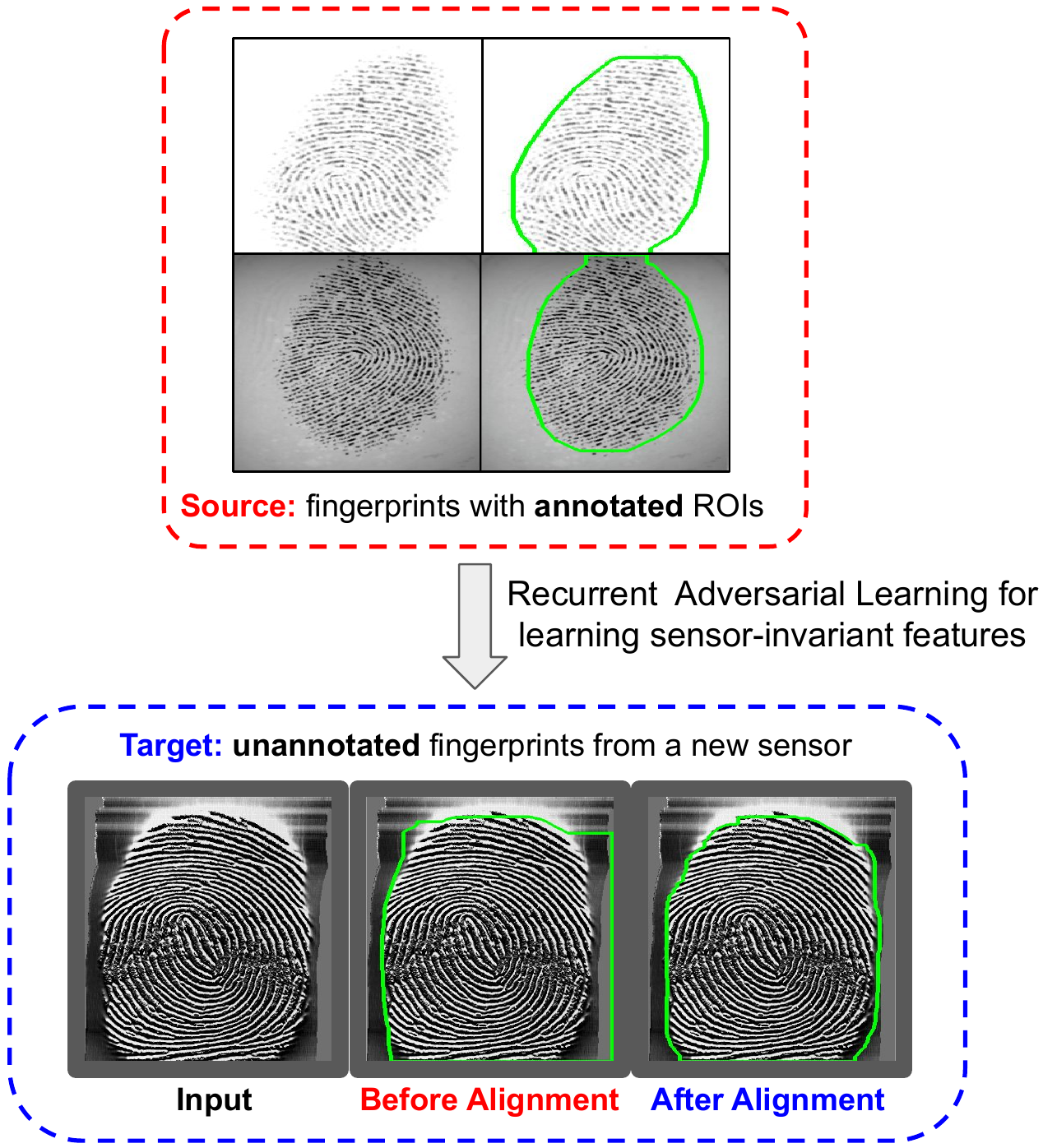}
\caption{Schematic diagram showcasing the benefit of proposed feature alignment method in improving segmentation performance on a sensor whose ground truth annotations are not available for training. For better understanding, roi marked fingerprints are presented instead of binary roi mask. }
\label{visualization_introduction}
\end{figure}

\par With the advent of Convolutional Neural Networks (CNNs) and its success in image processing applications, CNNs are now the state-of-the-art architectures for fingerprint processing, including roi segmentation. However, a limitation of CNNs is that the model trained on one domain often does not generalize well on other domains. In this research, we study this limitation of CNNs in the context of fingerprint roi segmentation. Specifically, this work is based on the hypothesis that the fingerprint roi segmentation performance is likely to be unsatisfactory on an unseen fingerprint sensor. In such a case, state-of-the-art fingerprint roi segmentation models need to be retrained using pixel-level manually marked roi corresponding to the newly introduced sensor.

\par Manually marking roi every time a new fingerprint sensor is introduced is impractical. Pixel-level annotation is highly expensive due to the human effort and necessary domain expertise. In order to save the cost and time required for the annotations required by the state-of-the-art, we approach the roi segmentation on a different sensor as an unsupervised domain adaptation problem. We propose a \textit{recurrent adversarial learning} framework to learn sensor-invariant features (see Figure \ref{visualization_introduction}) by aligning the features of training images (source) and fingerprint images corresponding to the target fingerprint sensor (without manually annotated roi).
\par The motivation for recurrent adversarial learning is derived from the human perception system as suggested in \cite{tu2009auto}. When humans observe a scene, their eye movements facilitate knowledge aggregation for scene understanding and also to refine perception \cite{schutz2011eye}. Similarly, recurrent adversarial learning enables proposed RA-RUnet (Recurrently Aligned RUnet) to iteratively refine features such that sensor-invariant features are learnt.

\section{Related Work}

\subsection{Fingerprint ROI Segmentation}
\subsubsection{Roi segmentation using classical image processing} Filtering in spatial and Fourier domain are one the most widely approaches for roi segmentation~\cite{seg3,seg10}. Morphological operations have also been explored for roi segmentation~\cite{seg6,seg12}. Some approaches exploit information originating from fingerprint ridge orientations to segment roi~\cite{seg8,seg9}.
\subsubsection{Roi segmentation using machine learning}
Initial learning based approaches propose clustering of image pixels to discriminate foreground from background \cite{seg4,seg7}. Handcrafted texture and intensity features have also been explored~\cite{seg11}. Recent fingerprint roi segmentation techniques exploit convolutional neural networks~\cite{seg1,Joshi_2021_WACV,seg2, joshi_data_uncertainty}.
\subsection{Interoperability Across Fingerprint Sensors}
Ross and Jain~\cite{ross2004biometric} conducted first ever study on cross-sensor matching performance and concluded poor interoperability. Initial approaches to improve interoperability include compensating for distortion~\cite{ross_nadgir} and resolution~\cite{han2006resolution, jang2007improving}. Texture features have shown promising performance~\cite{marasco,marasco2016improving,alshehri2018fingerprint,marasco2018enhancing}. Some approaches have exploited fusion based approaches~\cite{alshehri2018cross,alonso2006sensor}.

\par To summarize, these approaches primarily align minutiae to obtain invariance from either difference in resolution or texture across varying sensors. Different from these, this research is concentrated on improving interoperability of fingerprint roi segmentation without requiring manual annotations for the newly introduced sensors.
\subsection{Adversarial Learning}
Several studies showcase the success of adversarial learning framework in a variety of applications such as image generation~\cite{goodfellow2014generative,isola2017image}, audio-generation~\cite{kurmi2021collaborative}, domain adaptation~\cite{Kurmi_2021_WACV,ganin2015unsupervised,kurmi2019looking}, image in-painting~\cite{pathak2016context,yeh2017semantic}, incremental learning~\cite{Kurmi_2021_WACV_in} and fairness leaning. All of these approaches optimize the network with an adversarial discriminator. In the adversarial learning literature, some methods are proposed to improve the  discriminator for better learning. In~\cite{ghosh2018multi,pei2018multi}, multi discriminators are used to avoid the mode collapse problem faced during adversarial learning. A probabilistic discriminator is also explored for domain adaptation~\cite{saatchi2017bayesian,kurmi2019attending}. Similarly, a Unet based discriminator~\cite{schonfeld2020u} is used for image inpaining. In the fingerprints domain, adversarial learning is succesfully utilized for enhancement~\cite{xu2020augmentation, indu2019wacv}.
\subsection{Unsupervised Domain Adaptation}
The source and target features are generally aligned by minimizing the domain discrepancy using maximum mean discrepancy (MMD)~\cite{tzeng_arxiv2014}, correlation  alignment (CORAL)~\cite{sun_ECCV2016}, and adversarial metrics~\cite{ganin2015unsupervised,kurmi2019curriculum}. 
These discrepancy based methods are further extended to tackle the mode collapse problem~\cite{pei2018multi} and reduce negative transfer in domain adaptation. Generative adversarial networks~\cite{goodfellow2014generative} are well explored to generate labeled images from the target domain~\cite{li2020generating,sankaranarayanan2018generate}. Other than these approaches, uncertainty calibration~\cite{zheng2021rectifying,kurmi2019attending} and distributional matching~\cite{luo2020unsupervised} are also applied to adapt the classifier for the target domain.
\par Recently, affinity transformation based model ASA-Net~\cite{zhou2020affinity} is proposed to tackle the domain alignment problem in semantic segmentation. This model is based on the affinity relationship between adjacent pixels, termed as affinity space of source and target domain. A comparison between segmentation performance obtained by ASA-Net and proposed RA-RUnet is provided in Section \ref{sota}.
\begin{figure*}
		\centering \includegraphics[width=\textwidth,height=10cm]{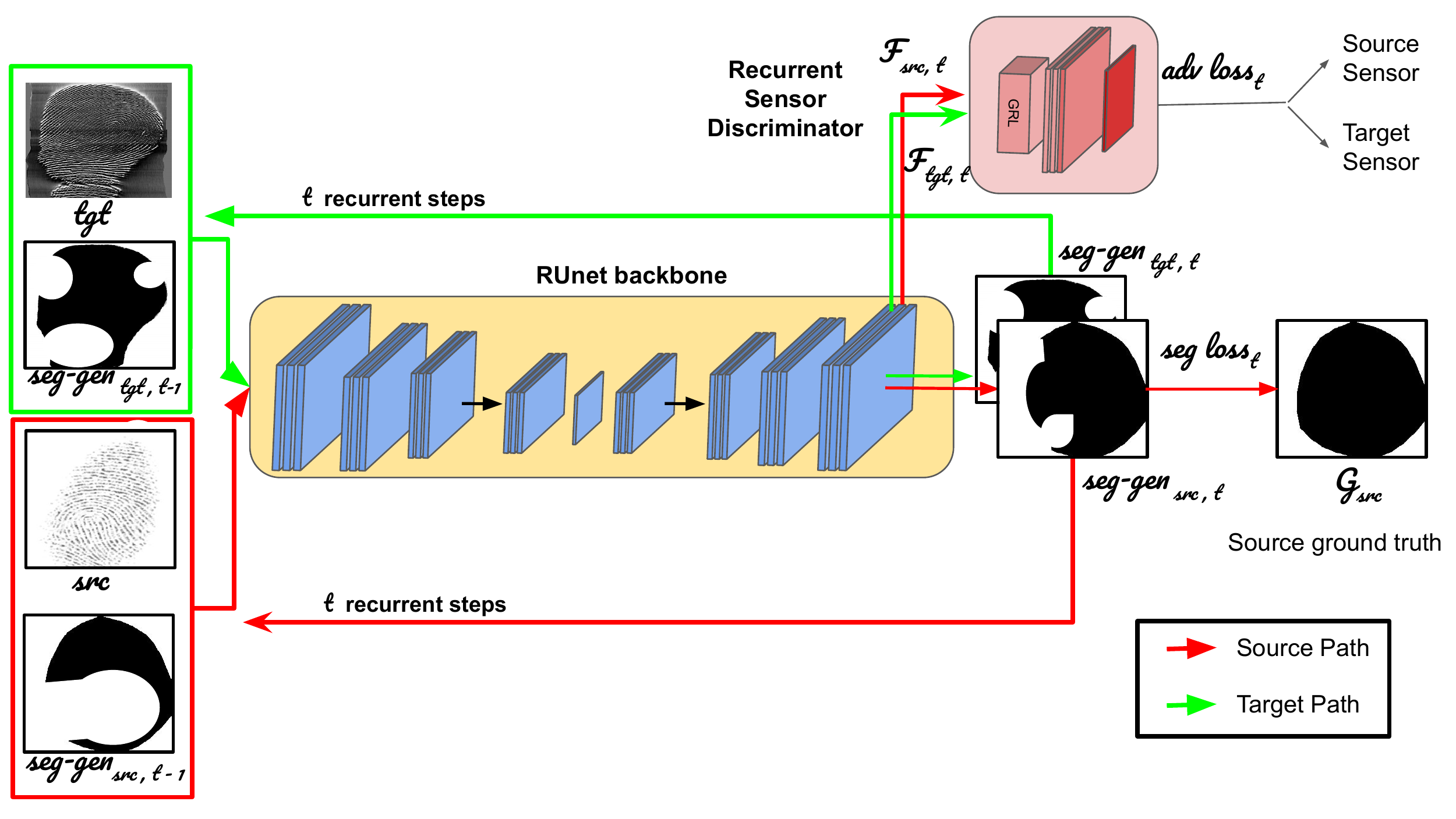}
		\caption{\textbf{Flowchart of proposed RA-RUnet}. RA-RUnet has two sub-networks: RUnet backbone and Recurrent Sensor Discriminator. While RUnet is used for segmentation, the latter is dedicated to feature alignment. For an input fingerprint image, Runet iteratively generates a roi segmentation mask, while the recurrent sensor discriminator classifies the features as derived from source or target sensor. The adversarial loss penalizes the discriminator if it misclassifies, whereas it penalizes RUnet if the discriminator makes a correct classification. This framework helps the backbone to learn sensor-invariant features. Segmentation loss is defined for the images from the source sensor such that the generated segmentation mask is close to the groundtruth. Please note that annotations of only source sensor are used for training the proposed RA-RUnet.}
		\label{flowchart_alignment}
\end{figure*}

\subsection{Research Contributions}
To the best of our knowledge, this research is the first work in fingerprints domain to exploit adversarial learning for learning sensor-invariant features. Experiments demonstrate the efficacy of proposed work in improving generalization ability of fingerprint roi segmentation on new and unseen sensors without relying on the corresponding manually marked roi mask.
\par We study the most challenging scenario in which features from synthetic fingerprints are aligned with features of real fingerprints. Real fingerprints used in this study are acquired from a variety of fingerprint sensors employing disparate sensing technology i.e. optical, capacitive and thermal. Visualizations of neural activations obtained from Seg-Grad-Cam~\cite{segcam} are also presented to provide insights on the improved segmentation performance.
\section{Algorithms Used For Benchmarking}
\label{benchmark}
To find the most suitable backbone architecture for this problem, we benchmark the following three state-of-the-art semantic segmentation models on fingerprint roi segmentation under the proposed setting.
\begin{itemize}

    \item \textbf{Unet:} Unet~\cite{ronneberger2015u} is a convolutional neural network that relies on an encoder-decoder network design, at its core. Unet is characterized by skip connections between encoding layer and decoding layers. These skip connections supply the required contextual information from neighbouring pixels and help the model to infer edge-level details. Owing to its compactness, Unet is one of the most effective network architecture for image processing applications involving small training datasets.
    
    \item \textbf{Ccnet:} Ccnet~\cite{huang2019ccnet} is a convolution neural network with a criss-cross attention unit. Its attention unit is designed to combine contextual information from vertical and horizontal directions with a limited increase in the computational requirements. Moreover, it obtains global information from the image via a recurrent mechanism. As a result, it learns contextual information from the image and obtains state-of-the-art segmentation performance.
    
    \item \textbf{Recurrent Unet (RUnet):} RUnet~\cite{wang2019recurrent} is a variant of Unet. It is especially crafted for applications with small amount of training dataset. Consequently, it is highly suitable for fingerprint roi segmentation due to the small size of training dataset. To maintain compactness, RUnet introduces recurrent units into the baseline Unet architecture. These recurrent units iteratively refine both the internal state of the RUnet and the segmentation mask. This improves the segmentation performance of baseline Unet architecture. Recurrent structure promotes parameters sharing and therefore, helps to avoid overfitting by limiting the number of parameters.
\end{itemize}
\section{Proposed Model}

 As shown in Figure \ref{flowchart_alignment}, proposed RA-RUnet consists of two networks: a segmentation network (RUnet backbone) and a discriminator network. Given an input fingerprint image, the segmentation network recurrently generates a segmented roi map whereas the discriminator is dedicated to iteratively classify whether the input features correspond to fingerprints from the source or target sensor.
 
\par Benchmarking results presented in Section \ref{benchmarking} reveal that RUnet is the most effective baseline network for segmentation. Therefore, proposed RA-RUnet employs RUnet as segmentation network and utilizes it recurrent architecture to improve feature alignment. Our contribution lies in introducing an adversarially trained recurrent sensor discriminator. Recurrent sensor discriminator ($\mathcal{D}$) iteratively classifies whether the recurrently generated feature maps of RUnet backbone network correspond to image from source or target sensor. This in effect, helps RUnet to learn sensor-invariant features.

\par To formalize, let us assume $src$ and $tgt$ denote input fingerprints acquired using source and target sensors respectively. For an iteration $t$, fingerprint image ($x_{src}$ or $x_{tgt}$) is concatenated with the segmentation mask ($seg\_gen_{src, t-1}$ or $seg\_gen_{tgt, t-1}$ respectively) generated at iteration ${t-1}$ and forwarded through the network. The network is then optimized by a combination of weighted segmentation and adversarial loss. The relationship between $seg\_gen_{tgt, t-1}$ and $seg\_gen_{tgt, t}$ is the same as defined in the baseline RUnet~\cite{wang2019recurrent} except a modification that now the optimization function also comprises of adversarial loss. Please note that only the ground truth roi mask of source, denoted by $g_{src}$ are used to train the RA-RUnet. The segmentation and adversarial loss are defined next.
\subsection{Segmentation Loss}
Segmentation loss is defined as the cross entropy loss between the ground truth roi mask and the segmentation mask generated by RA-RUnet. This loss guides the model to generate segmentation mask close to the ground truth. As this study assumes that ground truth roi masks of target are not available for training, segmentation losss is only defined for the fingerprints originated from source sensor.
Assume that $G_f$ is the feature extractor network of backbone architecture, parameterised by $\theta_f$. The source and target feature representation obtained at iteration $t$  are defined as:
\begin{equation}
   f_{src,t}=G_f(seg\_gen_{src, t-1},\theta_f) 
\end{equation}
\begin{equation}
   f_{tgt,t}=G_f(seg\_gen_{tgt, t-1},\theta_f) 
\end{equation}
The predicted mask for the source data is obtained by 
\begin{equation}
    seg\_gen_{src, t}= S_g( f_{src,t},\theta_s)
\end{equation}
where $\theta_s$ denotes the parameters of segmentation network ($S_g$).
The segmentation loss is defined as: 
\begin{equation}
	\begin{split}
seg \hspace{1 mm}loss_{t}=\frac{1}{n}\sum_{x=1}^{n} \mathcal{L}(seg\_gen_{src, x, t},g_{src, x})
	\end{split}
\end{equation}

 where $g_{src, x}$ denotes the $x^{th}$ pixel of the ground truth segmentation mask. $gen_{src, x, t}$ denotes the $x^{th}$ pixel of the segmentation mask generated for input fingerprint from source sensor, at iteration $t$. $\mathcal{L}$ and $n$ denote cross-entropy loss and the total number of pixels respectively.
 
\subsection{Adversarial Loss}
The adversarial loss is defined such that the backbone RUnet network is trained to minimize it while the discriminator ($\mathcal{D}$) is trained to maximize it. This is done by introducing a gradient reversal layer (GRL)~\cite{ganin2015unsupervised} before the discriminator. The gradient reversal layer reverses the gradient, only during backpropagation. Due to GRL the backbone Runet is penalized if the features ($f_{src}$ or $f_{tgt}$) are correctly classified by the discriminator. Due to the GRL based adversarial loss, RUnet learns the necessary sensor-invariant features for obtaining improved segmentation performance on the new target sensor.
The discriminator network ($\mathcal{D}$)  is trained by minimizing the sensor classification loss and is penalized if it misclassifies the domain of source and target fingerprint. Assuming $\theta_d$ denotes the parameters of discriminator network, the adversarial loss is defined as:
\begin{equation}
    	adv \hspace{1 mm}loss_{t}= \sum_{f\in  (f_{src,t} \cup f_{tgt,t})  } \mathcal{L}(\mathcal{D}(f,\theta_d),y)
\end{equation}

where $ y=0$ $\text{if $f \in f_{src,t} $}$ and $ y=1$ $\text{if $f \in  f_{tgt,t} $}$ .

\section{Training and Testing}
The objective function used to train the proposed RA-RUnet is defined as:
\begin{equation}
	\begin{split}
\mathcal{L}_{total}=\frac{1}{T}\sum_{t=1}^{T}\frac{1}{n}\sum_{x=1}^{n}[seg \hspace{1 mm}loss_{t}-\hspace{1 mm}\alpha \hspace{1 mm}adv \hspace{1 mm}loss_{t}]
	\end{split}
\end{equation}
where $\alpha$ denotes the weight of the adversarial loss and $T$ denotes the number of iterations. Please note that the negative sign in the loss function implies the adversarial relationship between the features extractor and the discriminator. We incorporate the gradient reversal layer~\cite{ganin2015unsupervised} to implement this in the proposed work. The loss is minimized using the standard backpropagation algorithm. While minimizing the total loss, the segmentation loss is minimized for the feature extractor, while adversarial loss is maximized. For the discriminator, the total loss is maximized; due to negative signs, the effective loss is minimized to train the discriminator.
The over all model is optimized using the following objectives:
\begin{equation}
    (\hat{\theta}_f,\hat{\theta_s}) =\underset{\begin{subarray}{c}
  \theta_f,\theta_s
  \end{subarray}}{\text{arg min }} \mathcal{L}_{total}(\theta_f,\theta_s,\theta_d)
\end{equation}

\begin{equation}
    (\hat{\theta}_d) =\underset{\begin{subarray}{c}
  \theta_d
  \end{subarray}}{\text{arg max }} \mathcal{L}_{total}(\theta_f,\theta_s,\theta_d)
\end{equation}

In this research, we use $\alpha$=1 and $T$=3. Baseline RUnet, RA-RUnet and its variants, all are trained for 5000 iterations. After training the proposed RA-RUnet, discriminator is discarded and only a forward pass through the segmentation network (baseline RUnet) is required to obtain the segmentation mask. Analysis of hyper-parameter $\alpha$ is performed in Section \ref{alpha}.
\section{Databases}
All the experiments reported in this study are performed on the publicly available Fingerprint Verification Competition (FVC) databases. FVC databases have two subsets: subset A defined for testing while subset B defined for training. In this work, source domain consists of FVC 2000DB4, FVC 2002DB4 and FVC 2004DB4 (subset B). A total of 240 images are used as the source domain images. Target domain consists of FVC 2000DB1-DB3, FVC 2002DB1-DB3 and FVC 2004DB1-DB3 (subset A) leading to a total of 7200 fingerprint images. Source domain is used only during training of RA-RUnet while target domain (without any manual annotations) is used both during training and testing. Details about the FVC databases used in this study are presented in Table \ref{table_databases}. The manually marked roi masks are taken from \cite{seg6}\footnote{\protect {https://figshare.com/articles/dataset/Benchmark\_for\_Fingerprint\_\\Segmentation\_Performance\_Evaluation/1294209}}.

\begin{table}
	\centering	
	\caption{Characteristics Of FVC Databases.}
	\begin{tabular}{|p{1.5cm}|p{1.4cm}|p{1cm}|p{1.3cm}|}
		\hline
		\textbf{Database}&\textbf{Sensing Technology}&\textbf{Domain}&\textbf{Size}\\
		\hline\hline
		2000 DB1  & Optical & Target& 300$\times$300\\
		\hline
		2000 DB2  & Capacitive & Target& 256$\times$364\\
		\hline
		2000 DB3  & Optical & Target& 448$\times$478\\
		\hline
		2000 DB4  & Synthetic &Source& 240$\times$320\\
		\hline
		2002 DB1  & Optical &Target& 388$\times$374\\
		\hline
		2002 DB2  & Optical &Target& 296$\times$560\\
		\hline
		2002 DB3 & Capacitive &Target& 300$\times$300\\
		\hline
		2002 DB4 & Synthetic &Source & 288$\times$384\\
		\hline
		
		2004 DB1 &Optical &Target & 640$\times$480\\
		\hline
		2004 DB2 &Optical &Target& 328$\times$364\\
		\hline
		2004 DB3 &Thermal &Target& 300$\times$480\\
		\hline
		2004 DB4&Synthetic & Source& 288$\times$384\\
		\hline
	\end{tabular}
	\label{table_databases}
\end{table}

\section{Evaluation Metrics}
To quantify the segmentation performance, this research is evaluated on two standard metrics used by the literature on segmentation: Dice \cite{dice1945measures} and Jaccard score \cite{choi2010survey}.
\begin{equation}
	\begin{split}
	Dice= \frac{2 \times TP}{(TP+FP)+(TP+FN)}
	\end{split}
\end{equation}
\begin{equation}
	\begin{split}
	Jaccard= \frac{TP}{(TP+FP+FN)}
	\end{split}
\end{equation}
where TP, TN, FP and FN denote true positive, true negative, false positive and false negative with respect to groundtruth roi mask.

\section{Results and Discussions}
\subsection{Benchmarking Results} \label{benchmarking}
Table \ref{table_dice_benchmark} and Table \ref{table_jaccard_benchmark} report the dice and jaccard similarity scores respectively, obtained by the benchmarked algorithms. Although Unet outperforms other architectures on five databases out of nine, however, it performs significantly bad on some. RUnet on the other hand, has the best generalization ability. As the proposed research is intended towards providing good generalization ability, we choose RUnet as the backbone architecture.
\par To give readers a perspective on how challenging the problem is, Table \ref{table_runet_synthetic_vs_runet_full} reports the dice and jaccard scores with and without using the annotations of roi masks of target sensor images. RUnet (synthetic) represents the scenario when roi masks from only synthetic images (source domain) are used. RUnet (full) on the other hand, represents the case when training is conducted in a fully supervised fashion, using the fingerprint images and the corresponding annotated roi masks from the training set (subset B) of both source and target domain \cite{Joshi_2021_WACV}. For comparison purposes, we have reported performance on only target sensor datasbases. As expected, the segmentation performance achieved by RUnet (synthetic) is significantly lower compared to RUnet (full). This compels us to propose aligning features of source and target domain to facilitate improved segmentation performance.
\begin{table}
	\centering	
	\caption{Dice score obtained by various state-of-the-art segmentation algorithms.}
	\begin{tabular}{|p{1.5cm}|p{1cm}|p{1.2cm}|p{1.2cm}|}
		\hline
		\textbf{Database}&\textbf{Unet ($\uparrow$)}&\textbf{Ccnet ($\uparrow$)}&\textbf{RUnet ($\uparrow$)}\\
		\hline\hline
		 2000DB1  &42.31&64.51&\textbf{71.24}\\
		2000DB2 &\textbf{76.62}&71.24&68.84\\
		2000DB3  &71.41&\textbf{91.58}&89.98\\
		2002DB1   &\textbf{97.79}&89.98&97.19\\
		2002DB2 &\textbf{92.77}&89.41&92.64\\
		2002DB3 &75.42&\textbf{91.80}&70.93\\
		2004DB1 &\textbf{98.86}&95.96&98.41\\
		2004DB2 &\textbf{92.50}&88.88&87.02\\
		2004DB3 &78.41&91.86&\textbf{94.36}\\
		\hline
	\end{tabular}
	\label{table_dice_benchmark}
\end{table}
\begin{table}
	\centering
	\caption{Jaccard score obtained by various state-of-the-art segmentation algorithms.}
\begin{tabular}{|p{1.5cm}|p{1cm}|p{1.2cm}|p{1.2cm}|}
		\hline
		\textbf{Database}&\textbf{Unet ($\uparrow$)}&\textbf{Ccnet ($\uparrow$)}&\textbf{RUnet ($\uparrow$)}\\
		\hline\hline
		 2000DB1  &27.73&53.41&\textbf{57.39}\\
		2000DB2 &\textbf{65.23}&59.12&55.90\\
		2000DB3  &57.76&\textbf{85.47}&83.20\\
		2002DB1   &\textbf{95.70}&82.92&94.57\\
		2002DB2 &87.09&81.95&\textbf{87.29}\\
		2002DB3 &61.79&\textbf{85.36}&58.03\\
		2004DB1 &\textbf{97.76}&92.50&96.88\\
		2004DB2 &\textbf{86.32}&81.17&79.15\\
		2004DB3 &65.59&85.47&\textbf{89.67}\\
		\hline
	\end{tabular}
	\label{table_jaccard_benchmark}
\end{table}
\begin{table}[]
\caption{Performance of RUnet when trained on only synthetic fingerprints Versus full training set.}
\centering
\begin{tabular}{|p{1cm}||p{1.3cm}|p{0.7cm}||p{1.3cm}|p{0.7cm}|}
\hline
\multirow{2}{*}{\textbf{Database}} & \multicolumn{2}{l|}{\textbf{Dice Score ($\uparrow$) }} & \multicolumn{2}{l|}{\textbf{Jaccard Similarity ($\uparrow$) }} \\ \cline{2-5} 
               &     \textbf{RUnet (synthetic)}     &     \textbf{RUnet (full)}       & \textbf{RUnet (synthetic)}          &       \textbf{RUnet (full)}     \\ \hline
2000DB1  &71.24& \textbf{93.34} &57.39& \textbf{88.15}  \\
2000DB2 &68.84& \textbf{92.39}  &55.90& \textbf{86.40} \\
2000DB3  &89.98& \textbf{96.50} &83.20& \textbf{93.74} \\
2002DB1 &97.19& \textbf{98.44} &94.57& \textbf{96.95} \\
2002DB2   &92.64&\textbf{97.28} &87.29 & \textbf{94.88} \\
2002DB3   &70.93& \textbf{95.53} &58.03& \textbf{91.83}  \\
2004DB1  &98.41& \textbf{99.38} &96.88& \textbf{98.78} \\
2004DB2   &87.02& \textbf{96.69} &79.15& \textbf{93.94}  \\
2004DB3   &94.36& \textbf{97.17} &89.67& \textbf{94.62}  \\
\hline
\end{tabular}
\label{table_runet_synthetic_vs_runet_full}
\end{table}
\begin{table}[]
\caption{Comparison Of Dice Score And Jaccard Similarity Obtained By Baseline RUnet And Proposed RA-RUnet.}
\centering
\begin{tabular}{|c||c|c||c|c|}
\hline
\multirow{2}{*}{\textbf{Database}} & \multicolumn{2}{l|}{\textbf{Dice Score ($\uparrow$) }} & \multicolumn{2}{l|}{\textbf{Jaccard Similarity ($\uparrow$) }} \\ \cline{2-5} 
               &     \textbf{RUnet}     &     \textbf{RA-RUnet}       & \textbf{RUnet}          &       \textbf{RA-RUnet}     \\ \hline
2000DB1  &71.24& \textbf{76.85} &57.39& \textbf{63.83}  \\
2000DB2 &68.84& \textbf{80.54}  &55.90& \textbf{69.10} \\
2000DB3  &89.98& \textbf{92.48} &83.20& \textbf{86.81} \\
2002DB1 &\textbf{97.19}& 95.77  &\textbf{94.57}& 91.96 \\
2002DB2   &\textbf{92.64}& 90.52 &\textbf{87.29} & 83.83 \\
2002DB3   &70.93& \textbf{75.41} &58.03& \textbf{61.69}  \\
2004DB1  &\textbf{98.41}&97.75  &\textbf{96.88}& 95.62 \\
2004DB2   &87.02& \textbf{93.78} &79.15& \textbf{88.68}  \\
2004DB3   &94.36& \textbf{94.50} &89.67& \textbf{89.72}  \\
\hline
\end{tabular}
\label{table_runet_vs_ra_runet}
\end{table}
\begin{figure*}
		\centering \includegraphics[width=0.82\textwidth,height=6.1cm]{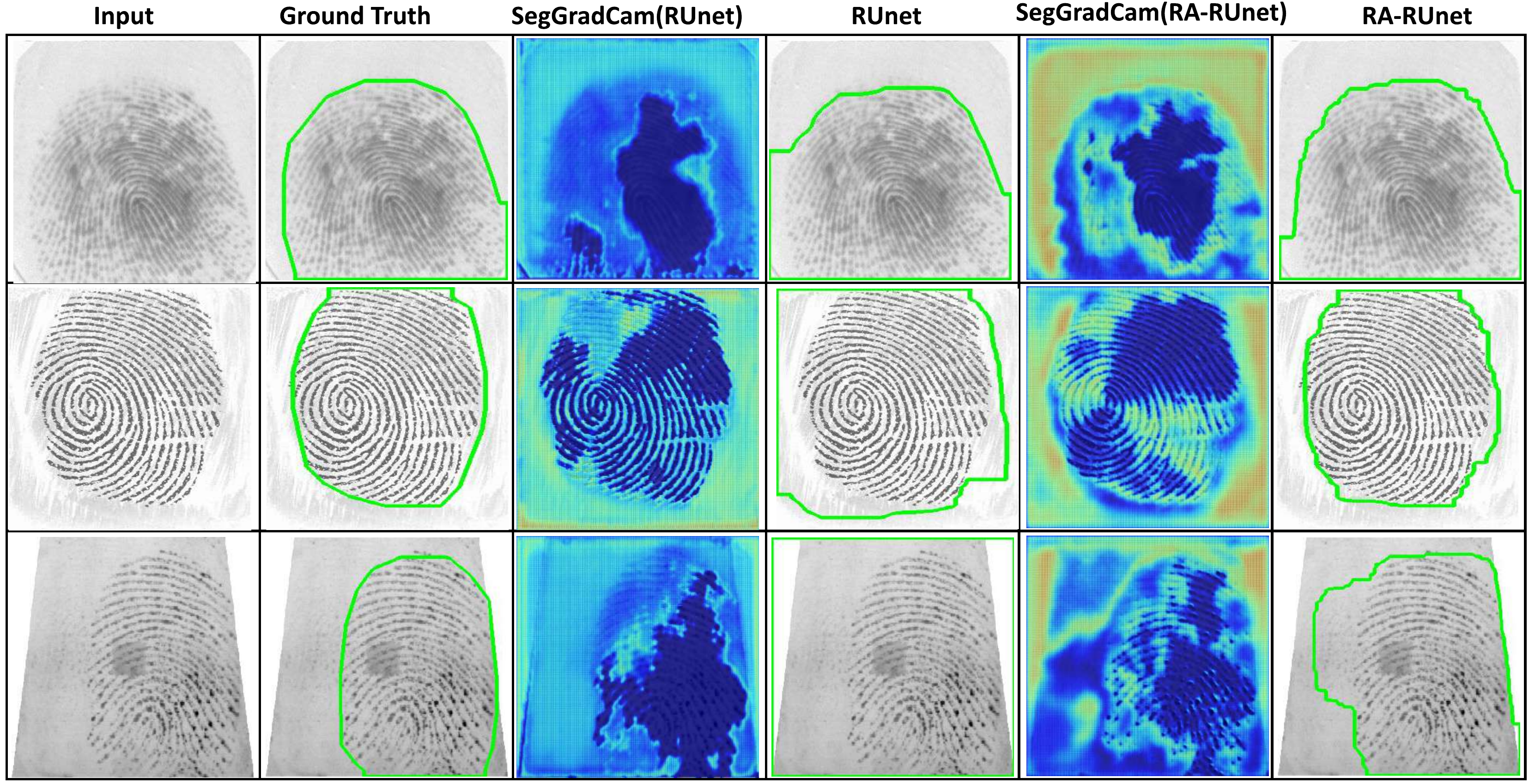}
		\caption{\textbf{Visualizations obtained using Seg-Grad-Cam} (best viewed in colour). Higher activations around boundaries are obtained by RA-RUnet compared to the baseline RUnet. This explains the improved generalization ability of RA-RUnet.}
		\label{segcam_visualization}
\end{figure*}

\subsection{Improved Generalization Ability}
Table \ref{table_runet_vs_ra_runet} compares segmentation performance of the proposed RA-RUnet with baseline RUnet. Visualizations obtained from Seg-Grad-Cam~\cite{segcam}, as shown in Figure \ref{segcam_visualization}, signify that higher activations around boundaries of foreground roi are obtained by RA-RUnet as compared to RUnet. As a result, RA-RUnet achieves better segmentation performance.
\subsection{Comparison With State-of-the-art} \label{sota}
Table \ref{table_sota_domain_adaptation} compares the segmentation performance of proposed RA-RUnet with the recently proposed ASA-Net~\cite{zhou2020affinity}, a state-of-the-art unsupervised domain adaptation method for semantic segmentation. Results reveal that the proposed RA-RUnet is better suited for fingerprints compared to ASA-Net as RA-RUnet significantly outperforms ASA-Net over eight out of nine databases.
\begin{table}[]
\caption{Comparison Of Dice Score And Jaccard Similarity Obtained By ASA-Net~\cite{zhou2020affinity} And Proposed RA-RUnet.}
\centering
\begin{tabular}{|c|c|c||c|c|}
\hline
\multirow{2}{*}{\textbf{Database}} & \multicolumn{2}{l|}{\textbf{Dice Score ($\uparrow$) }} & \multicolumn{2}{l|}{\textbf{Jaccard Similarity ($\uparrow$) }} \\ \cline{2-5} 
               &     \textbf{~\cite{zhou2020affinity}}     &     \textbf{RA-RUnet}       & \textbf{~\cite{zhou2020affinity}}          &       \textbf{RA-RUnet}     \\ \hline
2000DB1  &34.14& \textbf{76.85} &21.24& \textbf{63.83}  \\
2000DB2 &66.19& \textbf{80.54}  &52.63& \textbf{69.10} \\
2000DB3  &91.21& \textbf{92.48} &85.24& \textbf{86.81} \\
2002DB1 &91.00& \textbf{95.77}  &83.88& \textbf{91.96} \\
2002DB2   &88.73& \textbf{90.52} &81.18 & \textbf{83.83} \\
2002DB3   &\textbf{87.48}& 75.41 &\textbf{78.18}& 61.69  \\
2004DB1  &96.22&\textbf{97.75}  &92.82& \textbf{95.62} \\
2004DB2   &80.64& \textbf{93.78} &68.82& \textbf{88.68}  \\
2004DB3   &73.15& \textbf{94.50} &58.91& \textbf{89.72}  \\
\hline
\end{tabular}
\label{table_sota_domain_adaptation}
\end{table}
\begin{table}
	\centering	
	\caption{Dice score obtained by various state-of-the-art segmentation algorithms.}
	\begin{tabular}{|p{1cm}|p{1.2cm}|p{1.5cm}|p{1.5cm}|}
		\hline
		\textbf{Database}&\textbf{RUnet ($\uparrow$)}&\textbf{RA-RUnet ($T=$1) ($\uparrow$)}&\textbf{RA-RUnet ($T=$3)($\uparrow$)}\\
		\hline\hline
		 2000DB1  &71.24&70.52&\textbf{76.85}\\
		2000DB2 &68.84&78.21&\textbf{80.54}\\
		2000DB3  &89.98&\textbf{93.21}&92.48\\
		2002DB1   &\textbf{97.19}&95.72&\textit{95.77}\\
		2002DB2 &\textbf{92.64}&91.10&90.52\\
		2002DB3 &70.93&\textbf{88.35}&75.41\\
		2004DB1 &98.41&97.78&97.75\\
		2004DB2 &87.02&93.56&\textbf{93.78}\\
		2004DB3 &94.36&93.87&\textbf{94.50}\\
		\hline
	\end{tabular}
	\label{table_dice_single_versus_multi}
\end{table}
\begin{table}
	\centering
	\caption{Jaccard score obtained by various state-of-the-art segmentation algorithms.}
\begin{tabular}{|p{1cm}|p{1.2cm}|p{1.5cm}|p{1.5cm}|}
		\hline
		\textbf{Database}&\textbf{RUnet ($\uparrow$)}&\textbf{RA-RUnet ($T=$1) ($\uparrow$)}&\textbf{RA-RUnet ($T=$3)($\uparrow$)}\\
		\hline\hline
		 2000DB1  &57.39&56.15&\textbf{63.83}\\
		2000DB2 &55.90&66.22&\textbf{69.10}\\
		2000DB3  &83.20&\textbf{87.88}&86.81\\
		2002DB1   &\textbf{94.57}&91.87&\textit{91.96}\\
		2002DB2 &\textbf{87.29}&84.67&83.83\\
		2002DB3 &58.03&\textbf{79.84}&61.69\\
		2004DB1 &\textbf{96.88}&95.67&95.62\\
		2004DB2 &79.15&88.33&\textbf{88.68}\\
		2004DB3 &89.67&88.56&\textbf{89.72}\\
		\hline
	\end{tabular}
	\label{table_jaccard_single_versus_multi}
\end{table}
\begin{table*}[!]
\caption{Comparison Of Dice Score And Jaccard Similarity Obtained By Proposed RA-RUnet For Different Values Of $\alpha$.}
\centering
\begin{tabular}{|c|c|c|c|c|c||c|c|c|c|c|}
\hline
\multirow{2}{*}{\textbf{Database}} & \multicolumn{5}{c|}{\textbf{Dice Score ($\uparrow$) }} & \multicolumn{5}{c|}{\textbf{Jaccard Similarity ($\uparrow$) }} \\ \cline{2-11} 
               &     \textbf{0.2} &     \textbf{0.5}   &     \textbf{1} &     \textbf{2}&     \textbf{5} &     \textbf{0.2} &     \textbf{0.5}   &     \textbf{1} &     \textbf{2}&     \textbf{5}      \\ \hline
2000DB1  &42.98& 62.97 &\textbf{76.85}& 51.49 &47.51& 28.59 &47.50& \textbf{63.83} &36.16& 32.59  \\
2000DB2 &74.72& 78.35  &\textbf{80.54}& 76.14&76.93& 61.58 &66.02& \textbf{69.10} &64.03& 64.15  \\
2000DB3  &90.71& 92.74 &92.48& \textbf{92.92} &88.20& 83.89 &87.18& 86.81 &\textbf{87.42}& 80.54 \\
2002DB1 &95.27& 95.92 &95.77& \textbf{95.95}&94.89& 91.05 &92.23& 91.96 &\textbf{92.28}& 90.38  \\
2002DB2   &88.97&\textbf{ 90.66} &90.52 & 90.51&88.57& 81.29 &\textbf{84.03}& 83.83 &83.75& 80.82  \\
2002DB3   &\textbf{79.49}& 74.40 &75.41& 76.28&64.66& \textbf{66.91} &60.38& 61.69 &62.83& 49.00   \\
2004DB1  &97.54&97.84 &97.75& \textbf{97.90} &97.40& 95.23 &95.80& \textbf{95.92} &95.91& 94.97 \\
2004DB2   &90.10& 92.64 &\textbf{93.78}& 87.60 &79.89& 82.77 &86.93& \textbf{88.68} &79.77& 68.48  \\
2004DB3   &93.64& \textbf{94.94} &94.50& 93.76 &93.96& 88.18 &\textbf{90.48}& 89.72 &88.39& 88.73   \\
\hline
\end{tabular}
\label{table_parameter_alpha}
\end{table*}

\subsection{Recurrent Discrimination Promotes Improved Segmentation}
To showcase that recurrent discrimination yields improved segmentation, Table \ref{table_dice_single_versus_multi} compares the segmentation performance when the discrimination between source and target features is performed for just one iteration (T=1) compared to three (T=3). Please note that for both the cases, segmentation mask is iteratively refined thrice as done in baseline RUnet. Here $T$ characterizes no. of iterations used by the discriminator. For both the cases, as compared to the baseline RUnet, segmentation performance has significantly improved. However, model with T=3 outperforms the one with T=1 on five databases. Thus, recurrent discrimination helps to improve performance without adding any more model parameters.

\subsection{Effect of Hyper-parameter $\alpha$}
\label{alpha}
The loss function is a weighted combination of segmentation and adversarial loss characterized by hyper-parameter $\alpha$. As reported in Table \ref{table_parameter_alpha}, we observe that initially, as $\alpha$ increases, contribution of adversarial loss increases. As a result, sensor-invariant features are learnt and the segmentation performance improves. However, beyond a point, increasing $\alpha$ leads to decrease in contribution of segmentation loss which negatively affects the segmentation performance. Therefore, a careful choice of $\alpha$ needs to be made in order to have a well balanced loss function and best segmentation performance.



\section{Conclusion and Future Work}
Presented research is the first work on improving fingerprint roi segmentation performance on a new sensor without requiring its corresponding manually marked roi. Towards this, benchmarking of three state-of-the-art semantic segmentation models is performed to find the best backbone network. Afterwards, recurrent adversarial learning based feature alignment is performed to ensure that fingerprints originating from source and target fingerprint sensors have similar features. Experiments and visualizations demonstrate that proposed feature alignment framework improves the activations and segmentation performance around boundaries. Additionally, comparison between single-level and recurrent adversarial learning based alignment is performed to give insights on the improved roi segmentation performance. In future, feature alignment can be explored for other modules of a fingerprint matching system. 
\section*{Acknowledgment}
Authors thank the OPAL infrastructure from Université Cote d'Azur for computational resources and French National Research Agency, ANR for partial support (grant agreement ANR-18-CE92-0024). I. Joshi is partially supported by the Raman-Charpak Fellowship 2019.

\bibliographystyle{ieee}
\bibliography{reference}

\end{document}